%% file: main.tex
\documentclass[10pt,twocolumn,letterpaper]{article}
\usepackage[pagenumbers]{iccv}

\input{preamble}
\definecolor{iccvblue}{rgb}{0.21,0.49,0.74}
\usepackage[pagebackref,breaklinks,colorlinks,allcolors=iccvblue]{hyperref}

\title{Memory-efficient Streaming VideoLLMs for Real-time \\ Procedural Video Understanding}

\author{
Dibyadip Chatterjee\textsuperscript{\rm1,\rm3$*$}\, 
Edoardo Remelli\textsuperscript{\rm1}\, 
Yale Song\textsuperscript{\rm2}\, 
Bugra Tekin\textsuperscript{\rm1}\, 
Abhay Mittal\textsuperscript{\rm1}\, 
Bharat Bhatnagar\textsuperscript{\rm1} \\ 
Necati Cihan Camgöz\textsuperscript{\rm1}\, 
Shreyas Hampali\textsuperscript{\rm1}\, 
Eric Sauser\textsuperscript{\rm1}\, 
Shugao Ma\textsuperscript{\rm1}\, 
Angela Yao\textsuperscript{\rm3}\, 
Fadime Sener\textsuperscript{\rm1}\vspace{0.2cm} \\
$^1$ Meta Reality Labs  \quad 
$^2$ FAIR, Meta \quad 
$^3$ National University of Singapore \\
{\hypersetup{urlcolor=brightpink}\small\texttt{\href{https://dibschat.github.io/ProVideLLM}{dibschat.github.io/ProVideLLM}}}
\vspace{-0.5cm}
}

\begin{document}

\twocolumn[{
    \renewcommand\twocolumn[1][]{#1}
    \maketitle
    \centering
    \includegraphics[width=\linewidth]{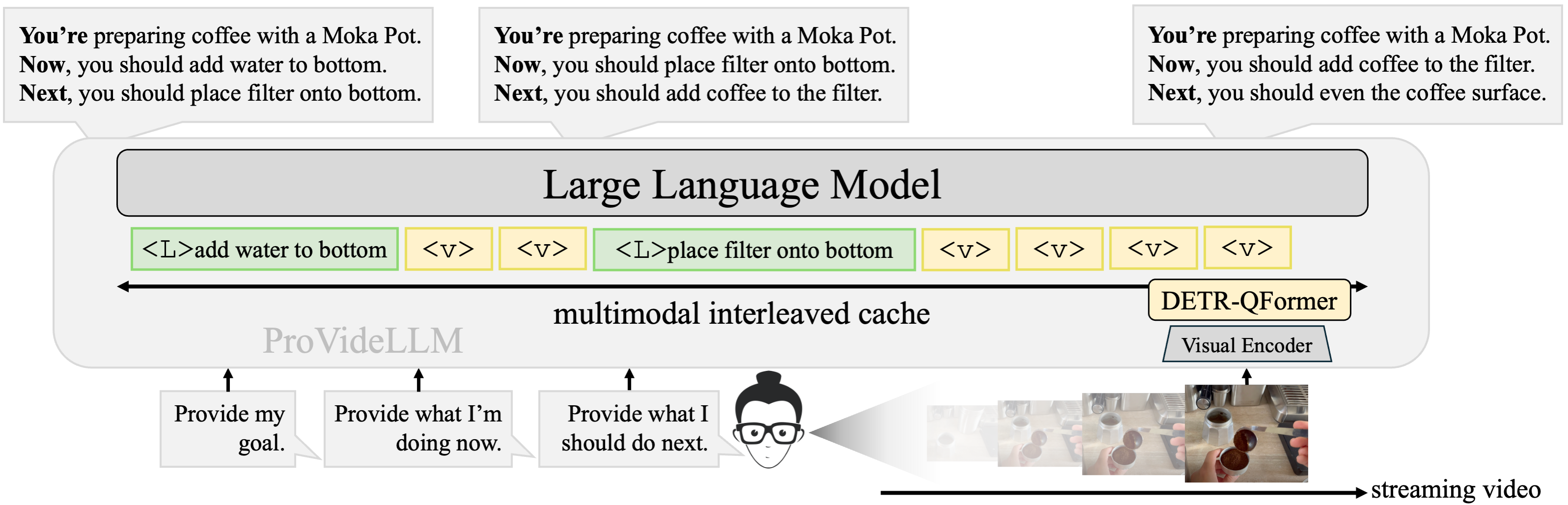}
    \captionof{figure}{\textbf{Overview of \mname{}.} A streaming video large language model for real-time procedural video tasks with a low memory footprint. It features a multimodal interleaved cache composed of textual descriptions of long-term observations spanning several minutes and visual tokens representing short-term observations spanning a few seconds. We also introduce a new DETR-QFormer connector for better fine-grained tokenization of the short-term. \mname{} is capable of handling multiple procedural tasks within a single model.}
    \label{fig:framework}
    \vspace{0.3cm}
}]

\blfootnote{\textsuperscript{$*$}Work done during an internship at Meta.}
%%%%%%%%%%%%%%%%%%%%%%%%%%%%%%%%%%%%%%%%%%%%%%%%%%%%%%%%%%%%%%%%%%%%%%%%

%%% Section: Abstract %%%%%%%%%%%%%%%%%%%%%%%%%%%%%%%%%%%%%%%%%%%%%%%%%
\begin{abstract}
We introduce \mname{}, an end-to-end framework for real-time procedural video understanding. \mname{} integrates a multimodal cache configured to store two types of tokens -- verbalized text tokens, which provide compressed textual summaries of long-term observations, and visual tokens, encoded with DETR-QFormer to capture fine-grained details from short-term observations. This design reduces token count by $22\times$ over existing methods in representing one hour of long-term observations while effectively encoding fine-granularity of the present. By interleaving these tokens in our multimodal cache, \mname{} ensures sub-linear scaling of memory and compute with video length, enabling per-frame streaming inference at 10 FPS and streaming dialogue at 25 FPS, with a minimal 2GB GPU memory footprint. 
\mname{} also sets new state-of-the-art results on six procedural tasks across four datasets.
\end{abstract}
%%%%%%%%%%%%%%%%%%%%%%%%%%%%%%%%%%%%%%%%%%%%%%%%%%%%%%%%%%%%%%%%%%%%%%%%

%%% Section: Introduction %%%%%%%%%%%%%%%%%%%%%%%%%%%%%%%%%%%%%%%%%%%%%%
\vspace{-0.5cm}
\section{Introduction}\label{sec:intro}
Procedural videos, spanning several minutes to hours, capture multi-step tasks~\cite{Wikihow} such as cooking~\cite{zhou2018towards, sener2019zero}, toy assembly~\cite{ragusa2021meccano, sener2022assembly101}, and furniture building~\cite{ben2021ikea, ben2024ikea}. With the growing adoption of AR assistants, streaming video understanding is essential for real-time procedural assistance. This requires real-time, online inference, as frames appear in a live video, without access to future observations. Despite its importance, streaming video understanding remains largely unexplored in existing procedural research~\cite{ding2023temporal, lai2024human, zhou2023procedure}.

Existing works on streaming video understanding~\cite{de2016online, shou2018online} focus on online action detection~\cite{xu2021long, zhao2022real, wang2023memory, pang2025context} or live conversations~\cite{chen2024videollm, wu2024videollm}. These use-cases target a small set of unrelated actions~\cite{de2016online,THUMOS14} or do not require long-term temporal modeling. In contrast, procedural videos are inherently long and causal, where each step depends on prior steps. Applying existing approaches to such videos significantly increases token count and memory overhead, leading to long-context degradation~\cite{hsieh2024ruler, liu2024lost} and making them unsuitable for streaming, even with excessive frame dropping~\cite{he2024malmm}. The semantic inter-connectivity of activity steps introduces redundancy when long-term observations are stored as raw frame tokens~\cite{xu2021long, wang2023memory}. Furthermore, while short-term observations are crucial for understanding the present, not all frame tokens are equally relevant, since procedural tasks often center around hand-object interactions~\cite{grauman2024ego, song2024ego4d, sener2022assembly101}.

In this paper, we propose an end-to-end framework for streaming procedural video understanding~(\cref{fig:framework}), addressing the above mentioned challenges of processing long-form videos. Specifically, we introduce \mname{}, a \underline{Pro}cedural \underline{Vide}o\underline{LLM} that comprises a visual (frame) encoder, a vision-language connector, and a pre-trained large language model (LLM)~\cite{radford2019language, achiam2023gpt, touvron2023llama, dubey2024llama} as the decoder. Our key contribution is a vision-language multimodal cache that efficiently manages streaming video data by balancing compression and representation quality of past observations.

We factorize the streaming past observations into long-term semantic history and short-term visual context. They are encoded as text and visual tokens, respectively, and interleaved in our multimodal cache. First, for long-term history spanning several minutes, \mname{} decoder predicts step labels per frame, groups similar steps into verbalized sequences (\eg~\textit{get $\rightarrow$ crack $\rightarrow$ whisk eggs}), and stores them in the cache. This semantic compression significantly reduces token count and also enchances current step prediction (\eg~\textit{pour mixture}).Second, to capture short-term visual context, typically spanning a few seconds, understanding the fine-grained changes across frames is crucial. However, standard VideoLLM encoders~\cite{radford2021learning, zhai2023sigmoid} face two key challenges in such modeling: \textit{(i)} temporal collapse, where low temporal variance conflates distinct actions and  \textit{(ii)} background distractors across frames~\cite{baradel2018object, rai2022transformed, chatterjee2023opening, zhang2022object}. We mitigate the former by using visual encoders trained without language alignment objectives~\cite{he2022masked, caron2021emerging, oquab2023dinov2}. We address the latter with a novel vision-text connector DETR-QFormer, pretrained to explicitly focus on hand-object interaction regions and compressing them into a smaller set of tokens.

Finally, in existing methods~\cite{xu2021long}, both long- and short-term tokens are visual and stored separately. Applying such separate caching to our multimodal tokens would require redundant attention recomputation when verbalizing short-term visual tokens into long-term text. To overcome this, we interleave text and visual tokens within our multimodal cache. This enables pre-computing attention, ensuring linear scaling of computation with cache size, which results in sub-linear scaling with video length, thereby significantly reducing latency for real-time performance.

In summary, our contributions are: \textit{(i)} a first-of-its-kind streaming multimodal cache with interleaved text and visual tokens, enabling real-time per-frame inference at 10 FPS and streaming dialogue at 25 FPS; \textit{(ii)} verbalizing long-term observations into compressed textual summaries and \textit{(iii)} modeling hand-object interactions for short-term observations using DETR-QFormer; all combined, it achieves a 22$\times$ reduction in token count compared to existing methods. \textit{(iv)} state-of-the-art results on six tasks across four datasets, making \mname{} the first real-time framework unifying recognition, anticipation, planning for procedural guidance.
\vspace{-0.2cm}
%%%%%%%%%%%%%%%%%%%%%%%%%%%%%%%%%%%%%%%%%%%%%%%%%%%%%%%%%%%%%%%%%%%%%%%%

%%% Section: Related Works %%%%%%%%%%%%%%%%%%%%%%%%%%%%%%%%%%%%%%%%%%%%%
\section{Related Works}\label{sec:related_works}
\textbf{Procedural Video Understanding} have primarily focused on instructional YouTube videos~\cite{zhou2018towards, tang2019coin, miech2019howto100m, zhukov2019cross}. Recent large-scale egocentric datasets~\cite{grauman2022ego4d, sener2022assembly101, song2024ego4d, grauman2024ego} have advanced research by providing longer videos with more activity steps. Existing works have focused on offline tasks~\cite{lin2022learning, zhou2023procedure, ding2023temporal}, where the entire video is available for prediction, as well as some online tasks, like procedural planning~\cite{chang2020procedure, islam2024propose,zhao2022p3iv}, forecasting~\cite{furnari2019would, sener2020temporal, girdhar2021anticipative, sener2019zero,narasimhan2022tl,zhao2024antgpt}, and mistake detection~\cite{sener2022assembly101, flaborea2024prego,ding2023every,lee2024error}. However, deploying these models for streaming inference results in high computational complexity where it grows quadratically with the video length. In contrast,~\mname{} incurs a sub-linear cost per update, both in terms of memory and compute, enabling efficient streaming inference.

\noindent\textbf{Video Large Language Models.} Multimodal large language models (MLLMs) have been popularized by BLIP-2-like~\cite{li2023blip, dai2023instructblip, zhu2023minigpt} and LLaVA-like~\cite{liu2024visual, liu2024improved, li2024llava} models where an LLM is trained to understand vision-text data. Extending these to video have led to the development of Video Large Language Models (VideoLLMs)~\cite{li2023videochat, maaz2023video, wu2024videollm, lin2023video, zhang2024video, zhao2024distilling}. However, most of these models are limited to offline tasks~\cite{tan2024koala, weng2024longvlm, wang2024videoagent, ren2024timechat} where the entire video is queried to answer a question. As such, architectural elements such as VideoQ-Former~\cite{zhang2023video} and temporal pooling layers~\cite{li2023videochat}, developed without causal requirements, limit streaming applications. Models like MA-LMM~\cite{he2024malmm} and VideoLLM-online~\cite{chen2024videollm} address video understanding \emph{online}, but are too computationally intensive to be deployed for efficient streaming inference on long-form videos. \mname{}~advances these models by introducing a multimodal interleaved cache for faster inference and proposing \cname{} for modeling fine-grained visual changes, improving accuracy.

\noindent\textbf{Streaming Video Understanding} has been explored for online action detection~\cite{de2016online, shou2018online, gao2019startnet} where dominant approaches~\cite{xu2021long, zhao2022real, wang2023memory, pang2025context} split streaming input into long- and short-term visual histories within an encoder-decoder architecture. However, storing both as visual tokens is redundant and computationally prohibitive for long-form procedural videos, where multiple actions are temporally connected. Instead, \mname{} verbalizes long-term observations into text which drastically decreases token count required to represent hour long videos, thus improving memory efficiency. 
%%%%%%%%%%%%%%%%%%%%%%%%%%%%%%%%%%%%%%%%%%%%%%%%%%%%%%%%%%%%%%%%%%%%%%%%

%%% Section: Method %%%%%%%%%%%%%%%%%%%%%%%%%%%%%%%%%%%%%%%%%%%%%%%%%%%%
\section{\mname}
\mname{} encodes long- and short-term streaming observations into text and visual tokens and interleaves them in a multimodal cache. We first present the preliminaries in~\cref{ssec:prelim}, followed by long- and short-term tokens generation in~\cref{ssec:long_past} and~\cref{ssec:short_past}. Finally, we present our multimodal cache in~\cref{ssec:streaming_cache}  and training objectives in~\cref{sec:training}.

\vspace{-0.05cm}
\subsection{Preliminaries}\label{ssec:prelim}
\vspace{-0.05cm}
\mname{} follows a LLaVA-style~\cite{liu2024visual} architecture, consisting of a visual encoder ($E_V$), a language decoder ($D_L$), and a vision-language connector ($C_{VL}$). Given a video $V = \{v_0, v_1, \dots, v_T\}$, the model's output ($o_t$) at any time step $t \in [0,T]$ is
\begin{equation}\small
 o_t = {\text \mname{}}(v_{0:t}) = D_L\left(\big\Vert_{i=0}^{t} C_{VL}(E_V(v_i))\right)
\end{equation}
where $\big\Vert$ denotes concatenation. While LSTR-inspired long short-term memory approaches~\cite{zhao2022real, wang2023memory} also use an encoder-decoder architecture, they rely exclusively on visual tokens~(\cref{fig:SIA}a.1). In contrast,~\mname{} incorporates a language decoder, which enables more efficient long-term compression, improves generalization, and supports multiple tasks within a single model.

\vspace{-0.05cm}
\subsection{Tokenizing long-term observations}\label{ssec:long_past}
\vspace{-0.05cm}
Procedural videos are inherently long-form, ranging from tens of minutes~\cite{sener2022assembly101} to several hours~\cite{song2024ego4d, grauman2024ego}. While modern LLMs~\cite{dubey2024llama, yang2024qwen2, team2024gemini} offer massive context lengths and, in theory, can process an hour-long raw video at 24 fps, performance degrades with larger context sizes~\cite{hsieh2024ruler, liu2024lost}, and training such models is highly memory-intensive~\cite{he2024malmm, zhang2024long}. Accurately recognizing ongoing steps require effectively encoding long-term context. Our key insight is that procedural steps in a video are temporally connected, \ie a sequence of step labels can serve as a strong predictor of ongoing steps. For instance, when ~\textit{``preparing an omelet''}, the step~\textit{``pour mixture into pan''} depends on earlier steps \textit{``get eggs from fridge, crack eggs into bowl, whisk eggs''}. Prior works on flow graphs~\cite{dvornik2022flow, NEURIPS2024_6d19163e} and planning~\cite{zhou2023procedure, ashutosh2023video, patel2023pretrained} also show that step sequences, often extracted from external sources~\cite{koupaee2018wikihow}, serve as strong baselines for procedural understanding.

\begin{figure}[t]
\centering 
\includegraphics[width=\linewidth]{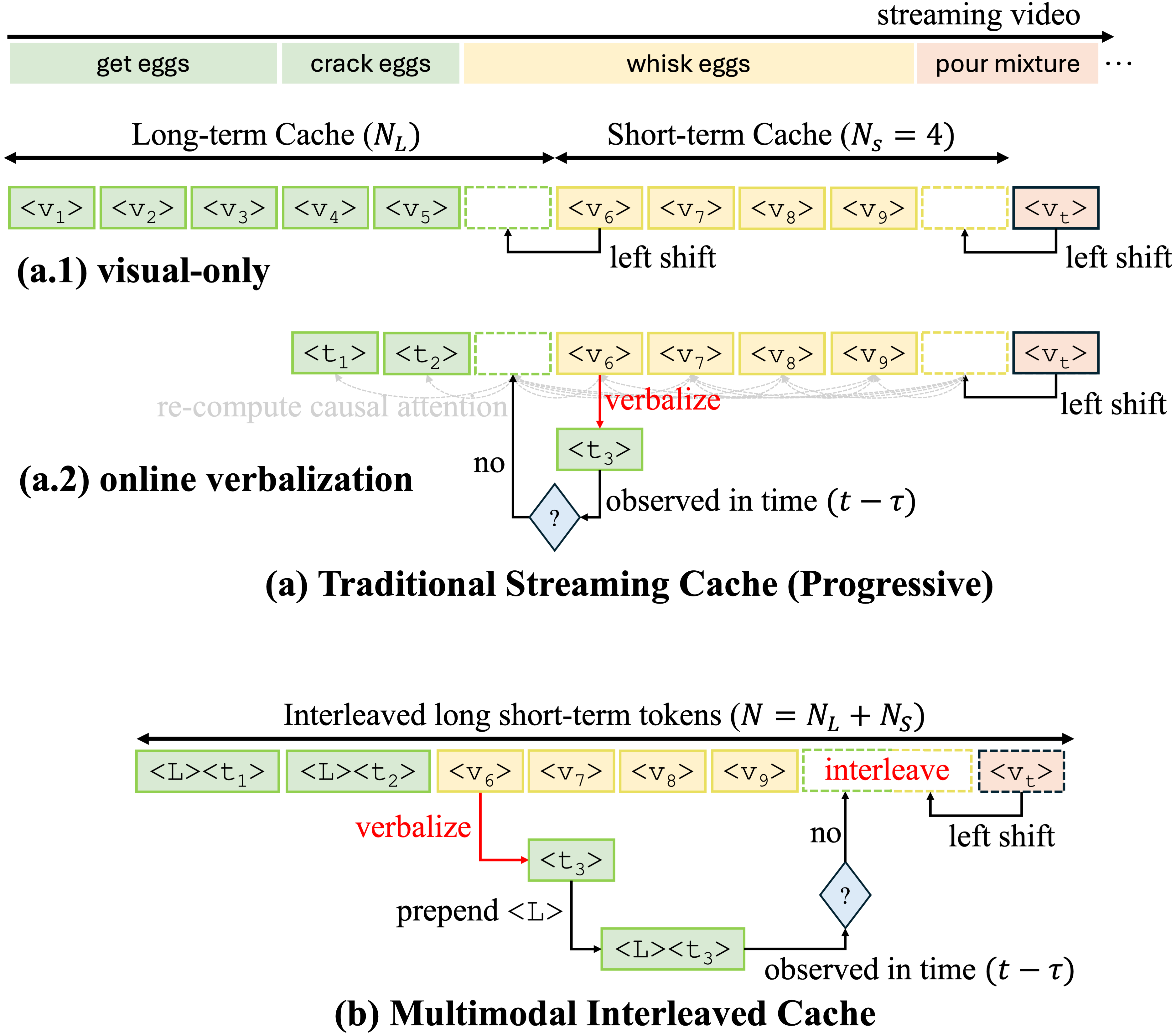}
\caption{\textbf{Converting short-term token to long-term under various caching strategies.} \textbf{(a)} Factorizing the streaming observations into long-term and short-term and ordering them progressively. 
\textbf{(a.1)} Conversion done at $O(1)$ cost but is unable to represent long-form videos due to massive token count and memory needs.
\textbf{(a.2)} Online verbalization can represent long-form videos at a manageable memory but is not suitable for streaming due to $O(N_S^2+N_L)$ conversion cost.
\textbf{(b)} Interleaving them reduces the conversion cost to $O(N)$, allowing streaming inference on long-form videos.}
\vspace{-0.2cm}
\label{fig:SIA}
\end{figure}

Instead of constructing computationally expensive graphs, we propose ``online verbalization''~(\cref{fig:SIA}a.2). Given streaming video, \mname{} decoder continuously generates action summaries (predictions) of the present observation on the fly, and groups semantically similar actions to verbalize the past~(\cref{fig:framework}). This approach efficiently compresses past visual frame tokens into information-dense language tokens. For example, in Ego4D Goal-Step~\cite{song2024ego4d}, storing one hour of long-term past requires only 630 verbalized text tokens on average. Compared to long short-term approaches~\cite{xu2021long, zhao2022real, wang2023memory}, our verbalization reduces token count by $22\times$ at the same sampling rate, significantly reducing the memory overhead.

\subsection{Tokenizing short-term observations}\label{ssec:short_past}
Our goal is to recognize procedural activity steps in a streaming manner. This requires understanding ``what is happening'' in the present while tracking ``what has happened in the past''. Language is a strong indicator of the semantic structure of step sequences, but determining if and when the step is performed requires fine-grained visual information. Therefore, we allocate more compute for understanding the short-term past, typically spanning 8-16 seconds, by keeping its visual tokens uncompressed. Standard VideoLLM encoders~\cite{liu2024visual, li2024llava} face two key challenges in capturing fine-grained short-term observations: \textit{(i)}~they generate temporally collapsed video features, a limitation we identify and address in~\cref{sssec::vis_enc}, and \textit{(ii)} they get biased towards high-level scene and background distractors, which we address in~\cref{sssec:detr_qformer}.
 
\begin{figure}[t]
\centering 
\includegraphics[width=0.99\linewidth]{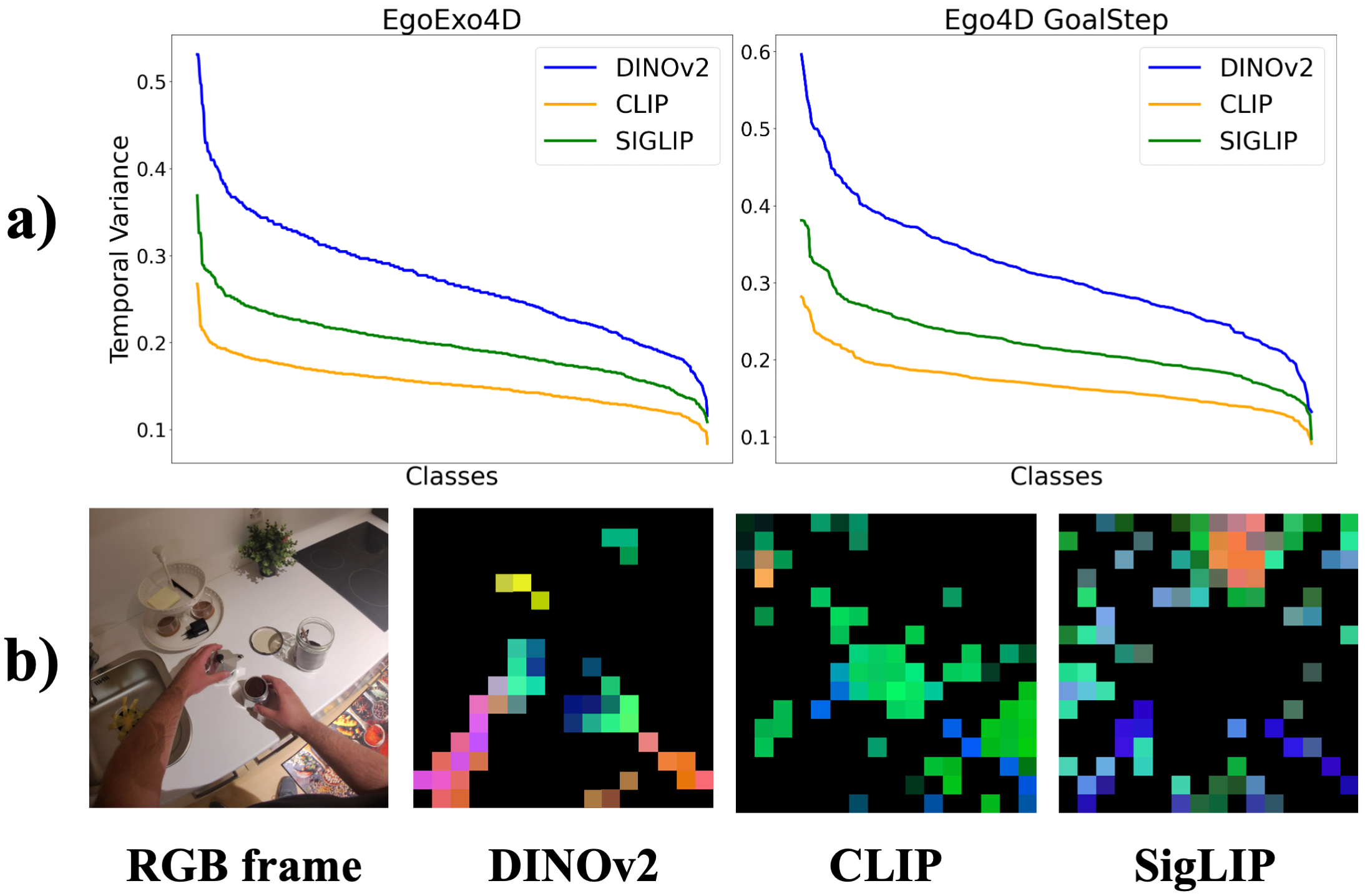}
\caption{
\textbf{a) Per-class mean temporal variance} on EgoExo4D \& Ego4D Goal-Step. Highly temporally varying classes on the left and low-variance classes on the right.
\textbf{b)~Visualization of the top-3 PCA components of patch tokens} computed across video frames, thresholded by first component. The primary factors of variation for DINOv2~\cite{oquab2023dinov2} are hands, and are unfocused and scattered across the image for language-aligned encoders, CLIP~\cite{radford2021learning} and SigLIP~\cite{zhai2023sigmoid}.
}
\vspace{-0.2cm}
\label{fig::variance_comparison}
\end{figure}

\subsubsection{Choosing the visual encoder}\label{sssec::vis_enc}
The predominant visual encoders for MLLMs are CLIP~\cite{radford2021learning} and SigLIP~\cite{zhai2023sigmoid, alabdulmohsin2024getting}. These encoders are pre-trained with a language-alignment objective for  integration with LLMs. However, because they are trained on high-level and sparse web-scale captions such as~\textit{``cooking in the kitchen''}, they struggle with fine-grained reasoning, particularly in answering the~\textit{``what?''} and~\textit{``how to?''} aspects of a video~\cite{xiao2024can, xiao2025videoqa}. Consequently, distinct images may yield high similarity scores in their feature space~\cite{tong2024eyes}, leading to low feature diversity across frames within a step. This results in relatively static frame features, a phenomemon we term ``temporal collapse". For certain fine-grained procedural steps, such as~\textit{``pick up" vs. ``put down"}, temporally collapsed features make it nearly impossible for the downstream LLM to decode the correct step.

\cref{fig::variance_comparison}a shows the impact of temporal collapse across three visual encoders -- two language-aligned ones~\cite{radford2021learning, zhai2023sigmoid} and one self-supervised\footnote{We chose DINOv2 due to its comparable scale to CLIP and SigLIP.}~\cite{oquab2023dinov2}. We plot temporal variance per class in descending order, with high-variance classes on the left and low-variance classes on the right (see Suppl.Sec.A for temporal variance computation). At both extremes, the trends are consistent across all encoders. High-variance classes, which often involve verbs such as~\textit{``get''},~\textit{``fetch''} or~\textit{``put''} correspond to longer videos with substantial scene and environmental changes. Conversely, low-variance classes involve adverbs such as~\textit{``slowly''} or ~\textit{``boil''} or repetitive actions like~\textit{``rotate''} or \textit{``peel''}, exhibit gradual state changes.  Among the three models, DINOv2 consistently captures temporal variations more effectively, covering a broader range of temporal variance across different classes. Hence, we choose DINOv2 as our visual encoder.

\begin{figure}[t]
\centering
\vspace{0.3cm}
\includegraphics[width=\linewidth]{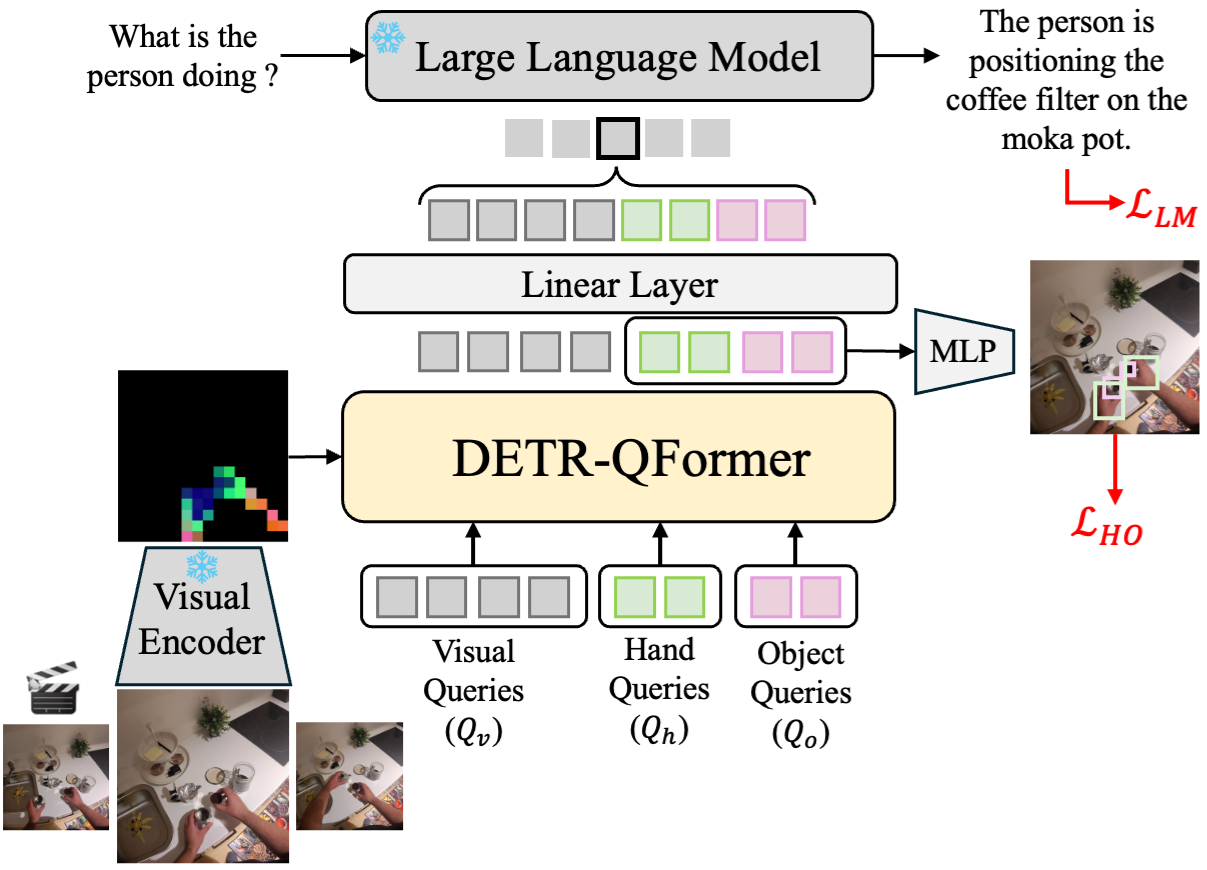}
\caption{\textbf{\cname{} architecture \& Stage-1 pre-training.} Trained to detect hands and objects-in-contact from pseudo ground truths bboxes generated by~\cite{shan2020understanding}, DETR-QFormer queries DINOv2 patch tokens to extract hand-object-focused activations.}
\vspace{-0.5cm}
\label{fig::connector}
\end{figure}

\subsubsection{Aligning hand-object transformations with text}\label{sssec:detr_qformer}
Procedural datasets predominantly capture hand-centric actions. For instance, nearly 99\% of actions in~\cite{song2024ego4d, grauman2024ego} involve hand interactions\footnote{Remaining 1\% are visually undetectable actions ``call for help, waiting.''}. Moreover, video frames often contain distracting background objects, while hand locations serve as strong cues for recognizing the ongoing actions~\cite{zhang2023helping, shamil2024utility}. Thus, a vision encoder focused on hand regions can enhance performance by mitigating background distractions and capturing fine-grained transformations more precisely.

Another contribution of our work is a vision-language connector, \cname{},  pretrained to attend to salient objects. In procedural videos, these are typically in contact with the hand. Interestingly, our choice of DINOv2 as the visual encoder provides an advantage as its patch tokens capture hands as primary factors of variation~\cref{fig::variance_comparison}b.  Leveraging this, we design \cname{} to query DINOv2 patch tokens, extracting hand-object activations and suppressing background distractions. \cname{} explicitly attends to hand-object activations using a DETR loss~\cite{carion2020end} and compresses salient regions into fewer tokens via QFormer's~\cite{li2023blip} cross-attention, enhancing representation quality. 

\noindent\textbf{Architecture.}
Our connector integrates Q-Former’s compression~\cite{li2023blip} and DETR’s bbox loss into a single cross-attention block. \cname{}, illustrated in~\cref{fig::connector}, is pretrained to detect the hands and objects-in-contact (\cref{sec:training}). The connector ($C_{VL}$) is a cross-attention transformer decoder with a set of $m$ learnable visual queries $Q_v = (q_{v_1}, q_{v_2},...q_{v_m})$. Key-value pairs ($K, V$) are generated from the flattened $16^2$ DINOv2 patch tokens using two projection heads ($W_K, W_V$) respectively. To explicitly model hand-object interactions, we append two sets of special queries -- hand queries $Q_h = (q_{h_1}, q_{h_2})$ for the left and right hands, and object queries $Q_o = (q_{o_1}, q_{o_2},\dots q_{o_k})$ corresponding to objects-in-contact. The role of visual queries $Q_v$ is to compress the $16^2\times D$ patch tokens into $m\times d$ tokens, where $m << 16^2$ and $d << D$. Whereas, outputs corresponding to the learnable hand-object queries ($Q^{\prime}_h, Q^{\prime}_o$) are passed through an MLP to make bounding box predictions, $\hat{B} = \{\hat{b}_{h_1}, \hat{b}_{h_2}, \hat{b}_{o_1}, \dots \hat{b}_{o_k}\}$.

\subsection{Multimodal Interleaved Cache}\label{ssec:streaming_cache}
The effectiveness of a procedural model is best assessed  when deployed for real-time inference. Na"ive long- and short-term approaches~\cite{xu2021long, wang2023memory} maintain separate caches for the two token types, arranged progressively with a single entry point (see~\cref{fig:SIA}a). With multimodal tokens, such setups require two distinct entry points, forcing attention recomputation over the short-term cache whenever a short-term token is verbalized into a long-term one. In the worst case, this leads to quadratic scaling of compute. To efficiently leverage online verbalization, we propose a ``multimodal interleaved cache'' that interleaves long-term text and short-term visual tokens from~\cref{ssec:long_past} and~\cref{ssec:short_past}.

\noindent\textbf{Interleaving long- and short-term tokens.}
We store the running short-term visual tokens and long-term text tokens in a single multimodal cache operating as a FIFO (First In, First Out) queue. It has one entry point and two distinct exits for short- and long-term tokens. Our entry and exit functions are detailed in~\cref{alg:streaming_cache_class}, where $N_S$ denotes the short-term cache length (number of frames) and $N_L$ the long-term cache length (number of observed steps). As frame tokens ($v_i$) exit the short-term span, the \mname{} decoder verbalizes them into long-term text tokens unless already cached within time $\tau$. To distinguish the two token types within the same cache, we prepend a special token (represented as \texttt{$<$L$>$}) to the long-term text tokens. \mname{} is instruction-tuned to recognize tokens associated with \texttt{$<$L$>$} as belonging to the long-term observations~(\cref{sec:sec::IT}). Since both long- and short-term tokens share a single entry point, visual and textual tokens are interleaved in the cache at any given time.

\begin{algorithm}[t]
\caption{Multimodal Interleaved Cache}
\label{alg:streaming_cache_class}
\scriptsize
\begin{algorithmic}[1]
 \State \texttt{\textcolor{keywordcolor}{class} MultimodalInterleavedCache:}
 \State \quad \texttt{\textcolor{keywordcolor}{def} \_\_init\_\_(self, N\_S, N\_L):}
 \State \quad \quad \texttt{self.tokens = []}
 \State \quad \quad \texttt{self.N\_S = N\_S}
 \State \quad \quad \texttt{self.N\_L = N\_L}
 \State \quad \texttt{\textcolor{commentcolor}{\# entry function: adds a new token to cache}}
 \State \quad \texttt{\textcolor{keywordcolor}{def} entry(self, token):}
 \State \quad \quad \texttt{self.tokens.append(token)}
 \State \quad \texttt{\textcolor{commentcolor}{\# exit function for short-term tokens}}
 \State \quad \texttt{\textcolor{keywordcolor}{def} exit\_short(self):}
 \State \quad \quad \texttt{\textcolor{keywordcolor}{if} self.tokens.count(\textcolor{functioncolor}{<v>}) $>$ self.N\_S:}
 \State \quad \quad \quad \texttt{self.tokens.pop(self.tokens.index(\textcolor{functioncolor}{<v>}))}
 \State \quad \texttt{\textcolor{commentcolor}{\# exit function for long-term tokens}}
 \State \quad \texttt{\textcolor{keywordcolor}{def} exit\_long(self):}
 \State \quad \quad \texttt{\textcolor{keywordcolor}{if} self.tokens.count(\textcolor{functioncolor}{<L>}) $>$ self.N\_L:}
 \State \quad \quad \quad \texttt{self.tokens.pop(self.tokens.index(\textcolor{functioncolor}{<L>}))}
\end{algorithmic}
\end{algorithm}

\noindent\textbf{Streaming inference.}
Every time a new token is added to the cache (\cref{alg:streaming_inference}), instead of re-computing attention across all tokens, we use the causal decoder's K-V cache~\cite{pope2023efficiently} to compute attention only for the new token over the previous $N-1$ tokens. This ensures that the computational complexity for processing a new frame scales linearly with cache size ($O(N)$). Additionally, maintaining a single interleaved cache enables direct application of optimized attention implementations~\cite{beltagy2020longformer, weng2024longvlm}, though such techniques remain orthogonal to our approach. More importantly, for a fixed $N_S$, compression via verbalization ensures that $N_L$ and consequently $N = (N_L + N_s)$ scales sub-linearly with the number of video frames. Hence, compute also scales sub-linearly with the number of streaming video frames. This makes our model well-suited for low-latency applications.

\begin{algorithm}[t]
\caption{Streaming Inference}
\label{alg:streaming_inference}
\scriptsize
\begin{algorithmic}[1]
 \State \texttt{\textcolor{keywordcolor}{def} inference(V, N\_S, N\_L, $\tau$):}
 \State \quad \texttt{cache = MultimodalInterleavedCache(N\_S, N\_L, $\tau$)}
 \State \quad \texttt{out = []}
 \State \quad \texttt{\textcolor{keywordcolor}{for} $v_i$ \textcolor{keywordcolor}{in} V:}
 \State \quad \quad \texttt{cache.entry($v_i$)} \textcolor{commentcolor}{\texttt{\# add frame tokens to cache}}
 \State \quad \quad \texttt{cache.exit\_short()} \textcolor{commentcolor}{\texttt{\# check short-term overflow}}
 \State \quad \quad \texttt{$t_i$ = \mname(cache.tokens)} \textcolor{commentcolor}{\texttt{\# process tokens}}
 \State \quad \quad \texttt{\textcolor{keywordcolor}{if} $t_i$ \textcolor{keywordcolor}{not in} out[-$\tau$:]:}
 \State \quad \quad \quad \texttt{cache.entry($t_i$)} \textcolor{commentcolor}{\texttt{\# add output to cache}}
 \State \quad \quad \quad \texttt{cache.exit\_long()} \textcolor{commentcolor}{\texttt{\# check long-term overflow}}
\end{algorithmic}
\end{algorithm}

\subsection{Training \mname}\label{sec:training}
We adopt a two-stage training approach similar to LLaVA -- (1) alignment pre-training to align the encoder's visual tokens to the decoder's language space, and (2) instruction tuning of the model for specific tasks.

\noindent\textbf{Stage-1: Alignment Pre-training.} 
In this stage, our connector \cname{} ($C_{VL}$) is trained while keeping the visual encoder and language decoder frozen~(\cref{fig::connector}). The objective is to caption short video clips featuring hand-centric actions using the standard language modeling loss~\cite{radford2018improving}:
 
\begin{equation}
\mathcal{L_{\text{LM}}}=-\frac{1}{|W|} \sum_{i=1}^{|W|} \log P\left(w_i \mid w_{<i}, V; C^{\Theta}_{VL}\right),
\end{equation}
where $V$ is the input video clip, $W = \{w_1, w_2, \cdots w_{|C|}\}$ is the ground-truth caption, and $C^{\Theta}_{VL}$ is our \cname{} parameterized by $\Theta$. For detecting hands and corresponding objects-in-contact, we use a DETR~\cite{carion2020end}-style Generalized IoU~\cite{rezatofighi2019generalized} loss:
\begin{equation}
 \mathcal{L}_{\text{HO}}(b_i, \hat{b}_{\sigma(i)}) = \mathcal{L}_{\text{iou}}(b_i, \hat{b}_{\sigma(i)}) + \|b_i - \hat{b}_{\sigma(i)}\|_1
\end{equation}
where $b_i$'s are the pseudo ground-truth bounding boxes of hands and objects-in-contact generated by an off-the-shelf detector~\cite{shan2020understanding} and $\hat{b}_{\sigma(i)} \in \hat{B}$ is the bounding box matched from the predicted set $\hat{B}$. For simplicity, we keep our loss category-agnostic.
The final loss for Stage-1 is thus:
\begin{equation}
 \mathcal{L} = \mathcal{L_{\text{LM}}} + \lambda_1\mathcal{L_{\text{HO}}}
\end{equation}
where $\lambda_1$ is a hyper-parameter. In addition to vision-language alignment, Stage-1 also enables our model to attend to hand-object interactions.

\noindent\textbf{Stage-2: Instruction Tuning.}\label{sec:sec::IT} 
This stage focuses on fine-tuning our model with task-specific instructions such as~\textit{“Provide the goal.”},~\textit{“Provide what I am doing now.”},~\textit{“Provide what I need to do next.”}, or~\textit{“Provide the next 10 steps I need to do.”}. For instruction tuning, we fine-tune~\cname{} and the language decoder using $\mathcal{L_{\text{LM}}}$, while keeping the visual encoder frozen.
For streaming tasks, we also train our model with interleaved special~\texttt{$<$L$>$} tokens to represent long-term cache entries, distinguishing them from standard prompts. See~Supp.Sec.5 for further details.
%%%%%%%%%%%%%%%%%%%%%%%%%%%%%%%%%%%%%%%%%%%%%%%%%%%%%%%%%%%%%%%%%%%%%%%%

%%% Section: Experiments %%%%%%%%%%%%%%%%%%%%%%%%%%%%%%%%%%%%%%%%%%%%%%%
\section{Experiments}
We conduct extensive evaluations of \mname~on six tasks across four datasets covering streaming~(\cref{sec:exp::streaming}), long-term~(\cref{sec:exp::forecast}) and fine-grained~(\cref{sec:exp::recognition}) procedural understanding. We further assess \mname{}'s performance in multi-task settings and its generalization to new datasets in~\cref{sec:exp::generalization}. Finally, we provide ablations in~\cref{sec:exp::ablations} and \mname{}'s runtime analysis in~\cref{sec:exp::runtime}.

\begin{table}[t] 
\centering
\setlength{\tabcolsep}{1.4mm}{
\resizebox{0.98\columnwidth}{!}{
\centering
\begin{tabular}{lcc}
\hline
\multirow{2}{*}{Model} & \multicolumn{2}{l}{per-frame mAP} \\ \cline{2-3} 
 & Val & Test \\ \hline
LSTR~\cite{xu2021long} (only short-term) & 8.8 & - \\
LSTR~\cite{xu2021long} & 8.9 & 8.1 \\
EgoOnly~\cite{wang2023ego} & 10.3 & 10.9 \\
\rowcolor{blue!5}\mname-1B/5 (no verbalization) & \underline{12.1} & \underline{12.2} \\
\rowcolor{blue!5}\mname-1B/5 (verbalization \& interleaving) & \textbf{13.0} & \textbf{12.9} \\ 
\hline
\end{tabular}
}}
\caption{\textbf{Ego4D}~\cite{song2024ego4d} \textbf{GoalStep Online Step Detection.} For fair comparison with LSTR, we consider 16s of short-term past and 128s of long-term past compressed into language tokens on avg.}
\label{tab:online_ego4dgoal}
\end{table}

\begin{table}[t] 
\centering
\setlength{\tabcolsep}{1.5mm}{
\resizebox{0.97\columnwidth}{!}{
\centering
\begin{tabular}{lcc }
\toprule
\textbf{Model} & \textbf{Views} & \textbf{Action} \\ \midrule
TempAgg~\cite{sener2020temporal} & 12 views & 8.5 \\
\quad $+$ Goal~\cite{zatsarynna2023action} & 12 views & \underline{12.0 }\\
\rowcolor{blue!5}\mname-8B/11~($+$long-term verbalization) & v4 & \textbf{13.8} \\ \bottomrule
\end{tabular}
}}
\caption{\textbf{Action Anticipation Results on Assembly101} validation set. Performance measured by class-mean Top-5 recall.}
\label{tab:online_a101}
\end{table}

\noindent\textbf{Implementation Details.}\label{sec:exp::implementation}
Depending on the choice of vision encoder and language decoder, we have two model variants. \mname-1B/5 is our efficiency-focused model designed for real-time streaming. It uses DINOv2 (ViT-S) with Llama-3.2-1B-Instruct~\cite{dubey2024llama} and requires 5 tokens per frame. \mname-8B/11 is our more performance-focused model using DINOv2 (ViT-L) and Llama-3.1-8B-Instruct and requires 11 tokens per frame. Furthermore, for benchmarking on youtube videos~\cite{tang2019coin}, we introduce an additional model \mname-8B/11+, which uses SigLIP's \texttt{[CLS]} token along with DINOv2 patch tokens. Unless otherwise mentioned we use $N_s = 64, N_l = 5$~\ie storing a maximum of 64 frames in the short-term cache and a maximum of 5 observed steps in the long-term cache. \cname{} is pre-trained (Stage-1) on EPIC-KITCHENS~\cite{damen2022rescaling} with pseudo ground-truth bounding boxes from 100DOH~\cite{shan2020understanding}. For Stage-2, the LLM (decoder) is fine-tuned using LoRA~\cite{hu2022lora} with $r$=128 and $\alpha$=256. Additional implementation details are provided in Supp.Sec.5.

\noindent\textbf{Datasets.}
We evaluate our model's performance on four large-scale procedural datasets -- EgoExo4D~\cite{grauman2024ego}, Ego4D-GoalStep~\cite{song2024ego4d}, COIN~\cite{tang2019coin}, and Assembly101~\cite{sener2022assembly101}. 
As shown in~\cref{sec:exp::main}, \mname{} sets state-of-the-art on both egocentric and YouTube benchmarks, demonstrating robustness across diverse video domains.

\subsection{Main results}\label{sec:exp::main}

\subsubsection{Online Step Detection}\label{sec:exp::streaming}
We report results in~\cref{tab:online_ego4dgoal}, using per-frame mAP on action classes (steps + substeps) at 4fps, a standard metric in online action detection~\cite{zhao2022real, wang2023memory}. We evaluate \mname-1B/5 with and without long-term verbalization in rows four and five, respectively. Results for LSTR~\cite{xu2021long} and EgoOnly~\cite{wang2023ego} are taken from~\cite{song2024ego4d}. Accordingly, LSTR spans $128s$ in long-term memory and $16s$ in short-term memory. In~\cite{song2024ego4d}, each step spans on average $32s$ with around $5.7$ tokens representing each step. Our long-term cache can encode $128s$ of past using only 22 tokens at $N_L = 4$ steps. To measure improvements from our long-term design, we set the cache length $N=86$ (spanning LSTR's temporal context), split into $(N_S=86, N_L=0)$ and $(N_S=64, N_L=4)$ in rows four and five, respectively. Our model surpasses both baselines, outperforming LSTR by over 4\% at similar temporal span. \mname{} also achieves greater gains from long-term context compared to LSTR, highlighting the effectiveness of verbalizing the long-term past.

\begin{table}[t] 
\centering
\setlength{\tabcolsep}{1.5mm}{
\resizebox{0.97\columnwidth}{!}{
\centering
\begin{tabular}{ lc }
\toprule
Method & Top-1 Acc. (\%) \\ \midrule
TimeSformer~\cite{bertasius2021space} & 34.0 \\
DistantSup & 39.4 \\
VideoTF & 42.4 \\
ProcedureVRL & 46.8 \\
VEDIT~\cite{lin2024vedit} & 51.8 \\ \midrule
VideoLLM-online~\cite{chen2024videollm} & 49.1 \\
VideoLLM-MoD~\cite{wu2024videollm} & 49.7 \\
\rowcolor{blue!5}\mname-8B/11+ & \underline{52.1} \\ 
\rowcolor{blue!5}\mname-8B/11+~($+$long-term verbalization) & \textbf{53.6} \\
\bottomrule
\end{tabular}
}} 
\caption{\textbf{Next Step Forecasting in COIN.} The last four rows including \mname~are VideoLLMs.}
\label{tab:coin_forecasting}
\end{table} 

\subsubsection{Online Step Forecasting}\label{sec:exp::forecast}
Step forecasting requires modeling long-term dependencies, making it an ideal task to evaluate our contributions. We evaluate \mname~for step forecasting on COIN and Assembly101 measured by Top-1 accuracy and Top-5 Recall~\cite{furnari2018leveraging}, respectively. Our model achieves state-of-the-art performance on both datasets as shown in \cref{tab:coin_forecasting} and \cref{tab:online_a101}. On Assembly101, we surpass~\cite{zatsarynna2023action} by 1.8\% using only a single view. On COIN, we use Llama-3-8B-Instruct to ensure fair comparisons with VideoLLMs~\cite{chen2024videollm, wu2024videollm} and achieve a performance boost of around 4\%. Moreover, while state-of-the-art models~\cite{chen2024videollm, wu2024videollm} use approximately 600 to 6000 visual tokens to model past context, our approach spans a larger temporal context with only 200 tokens through verbalization.

\subsubsection{Step Recognition}\label{sec:exp::recognition}

In~\cref{tab:offline_egoexo} and~\cref{tab:offline_coin}, we evaluate \mname~on fine-grained step recognition on EgoExo4D and COIN. We uniformly sample 16 frames from each segment to construct our short-term cache ($N_s$) and use the prediction from the last frame as our output. The closest model, VideoLLM-MoD~\cite{wu2024videollm}, uses 10 tokens per frame, but densely samples frames at 2fps, resulting in a larger token count. In both cases, we outperform VideoLLM-MoD by 1.74\% and 3.9\% respectively. Moreover, using only egocentric video, we achieve the best results on the EgoExo4D Fine-grained Recognition Challenge, with a 9.2\% improvement over the second-ranked model.

\begin{table}[t] 
\centering
\setlength{\tabcolsep}{1.5mm}{
\resizebox{0.95\columnwidth}{!}{
\centering
\begin{tabular}{ lcc }
\toprule
\textbf{Model} & \textbf{\begin{tabular}[c]{@{}c@{}}Training \\ Views\end{tabular}} & \textbf{\begin{tabular}[c]{@{}c@{}}Ego Acc. (\%) \\ Val / Test\end{tabular}} \\ \midrule
TimeSFormer~\cite{bertasius2021space} & ego & 35.13 / 35.93 \\
EgoVLPv2~\cite{pramanick2023egovlpv2} & ego+exo & 39.10 / 38.76 \\
Viewpoint Distillation & ego+exo & 38.19 / 39.49 \\
View Invariant Encoder~\cite{oord2018representation}\quad\quad\quad & ego+exo & 40.34 / \underline{41.53} \\ \midrule
VideoLLM-MoD\textsuperscript{$\dagger$}~\cite{wu2024videollm} & ego & \underline{42.62} / \quad --- \quad \\ 
\rowcolor{blue!5}\mname-8B/11 & ego & \textbf{44.36} / \textbf{50.74} \\ \bottomrule
\end{tabular}
}} 
\caption{\textbf{EgoExo4D Fine-grained Keystep Recognition.} Test results obtained from EgoExo4D's Fine-grained Keystep Recognition Challenge on EvalAI. \textsuperscript{$~\dagger$} Uses 2fps sampling per segment.}
\label{tab:offline_egoexo}
\end{table}

\begin{table}[t] 
\centering
\setlength{\tabcolsep}{8mm}{
\resizebox{0.95\columnwidth}{!}{
\centering
\begin{tabular}{lc}
\toprule
Method & Top-1 Accuracy (\%) \\ \midrule
TimeSformer~\cite{bertasius2021space} & 34.0 \\
Paprika~\cite{zhou2023procedure} & 51.0 \\
DistantSup~\cite{lin2022learning} & 54.1 \\
ProcedureVRL~\cite{zhong2023learning} & 56.9 \\
VideoTaskGraph~\cite{ashutosh2024video} & 57.2 \\
VEDIT~\cite{lin2024vedit} & 62.7 \\ \midrule
VideoLLM-online~\cite{chen2024videollm} & 63.1 \\
VideoLLM-MoD~\cite{wu2024videollm}\quad\quad\quad & \underline{63.4} \\
\rowcolor{blue!5}\mname-8B/11+ & \textbf{67.3} \\ \bottomrule
\end{tabular}
}}
\caption{\textbf{COIN Step Recognition.} Last three rows are VideoLLMs.}
\vspace{-3mm}
\label{tab:offline_coin}
\end{table}

\subsection{Generalized Procedural Understanding}\label{sec:exp::generalization}
In this section, we explore additional capabilities of \mname, leveraging the advantages of using a pretrained LLM as the language decoder. This section provides insights into developing a~\textit{``generalist''} model that advance procedural video understanding within a unified framework. We first evaluate \mname's ability to multi-task and then present our benchmark for cross dataset generalization.

\begin{table}[t] 
\centering
\setlength{\tabcolsep}{2mm}  
\resizebox{0.95\columnwidth}{!}{
\begin{tabular}{lccccc}
\toprule
\multirow{2}{*}{Model} & \multicolumn{5}{c}{COIN Benchmark Top-1 Accuracy (\%)} \\
 & Step & Task & Next & Proc. & Proc.+ \\ \midrule
VideoTF\textsuperscript{$\dagger$}~\cite{narasimhan2023learning} & 56.5 & 91.0 & 42.4 & 40.2 & 46.4 \\
VideoLLM-online~\cite{chen2024videollm} & 63.1 & 92.7 & 49.1 & 49.8 & \underline{54.1} \\ 
VideoLLM-MoD~\cite{wu2024videollm} & \underline{63.4} & \underline{92.8} & 49.7 & \underline{49.8} & 53.3 \\ 
\rowcolor{blue!5}\mname-8B/11+ & \textbf{66.9} & \textbf{95.0} & \textbf{50.5} & \textbf{51.0} & \textbf{55.9} \\ \bottomrule
\end{tabular}
}
\caption{\textbf{Multi-task learning on COIN.} Step -- step recognition, Task -- task recognition, Next -- next step forecasting, Proc. -- long-term forecasting and Proc+ -- long-term forecasting with task.~\textsuperscript{$\dagger$}VideoTF~\cite{narasimhan2023learning} is trained separately for each task.}
\label{tab:multitask}
\end{table} 

\begin{table}[h]
\centering
\setlength{\tabcolsep}{2mm}{
\resizebox{0.95\columnwidth}{!}{
\centering
\begin{tabular}{lcc }
\toprule
\textbf{Model} & \textbf{Ego Acc. (\%)} \\ \midrule
VideoLLaVA~\cite{lin2023video} + Llama-3.1-8B~\cite{dubey2024llama} & 3.6 \\
LLaVA-OneVision~\cite{li2024llava} + Llama-3.1-8B~\cite{dubey2024llama} & \underline{8.3} \\
\rowcolor{blue!5}\mname-8B/11 & \textbf{12.4}\\ \bottomrule
\end{tabular}
}}
\caption{\textbf{Cross-dataset generalization.} Evaluated on EgoExo4D KeyStep recognition. Our model, trained on Ego4D Goal-Step is compared with a mixture of vision + language experts.}
\label{tab:cross_dataset}
\end{table}

\noindent\textbf{Multi-task learning.}\label{sec:exp:multitask}
In~\cref{tab:multitask}, we present our results for multi-task learning, an underexplored aspect of procedural understanding. Existing models are trained only for isolated tasks, such as step recognition and forecasting. We compare our model with recent VideoLLMs, also trained in a multi-task setting, on 5 tasks. Despite using the same LLM as~\cite{chen2024videollm, wu2024videollm}, our model outperforms these baselines across all tasks. 

\noindent\textbf{Cross-Dataset Generalization.}
In~\cref{tab:cross_dataset}, we present a cross-dataset generalization benchmark for procedural videos using the EgoExo4D fine-grained keystep recognition task. 
We compare two recent VideoLLMs~\cite{lin2023video, li2024llava} which were primarily trained for video captioning and use Llama-3.1-8B-Instruct to convert the captions into EgoExo4D classes. Our baseline models act as a combination of visual and language experts. \mname, using the same LLM and trained on the much smaller Ego4D Goal-Step dataset, significantly outperforms these baselines. Despite these advances, performance remains modest, highlighting the need for further research in procedural generalization. Additional details are provided in the Supplemental Material.

\subsection{Ablations}\label{sec:exp::ablations}
\noindent\textbf{Visual Encoders \& DETR-QFormer.}
In~\cref{tab:ablation_hand-qformer}, we ablate visual encoders--SigLIP and DINOv2—and connectors--MLP, QFormer, and our DETR-QFormer--on EgoExo4D keystep recognition. First, comparing encoders across connectors, SigLIP’s CLS tokens, being language-aligned, outperform DINOv2; however, with more patch tokens, DINOv2 surpasses SigLIP, aligning with our observations in~\cref{fig::variance_comparison}. Second, comparing connectors across encoders, Q-Former boosts DINOv2's performance (+1.2\%, +1.9\%) but has minimal impact on SigLIP (+0.8\%, -0.1\%), highlighting DINOv2’s need for a stronger connector. Finally, \cname{} further improves DINOv2 (+1.1\%, +2.6\%), with gains increasing at higher resolutions but degrading SigLIP’s performance due to weak hand-object activations in its patch tokens. Notably,~\cname{} remains highly token-efficient with DINOv2, even when compressing to only hand-object queries, reinforcing its advantage for low-latency streaming.

\begin{figure}[t]
\centering 
\includegraphics[width=\linewidth]{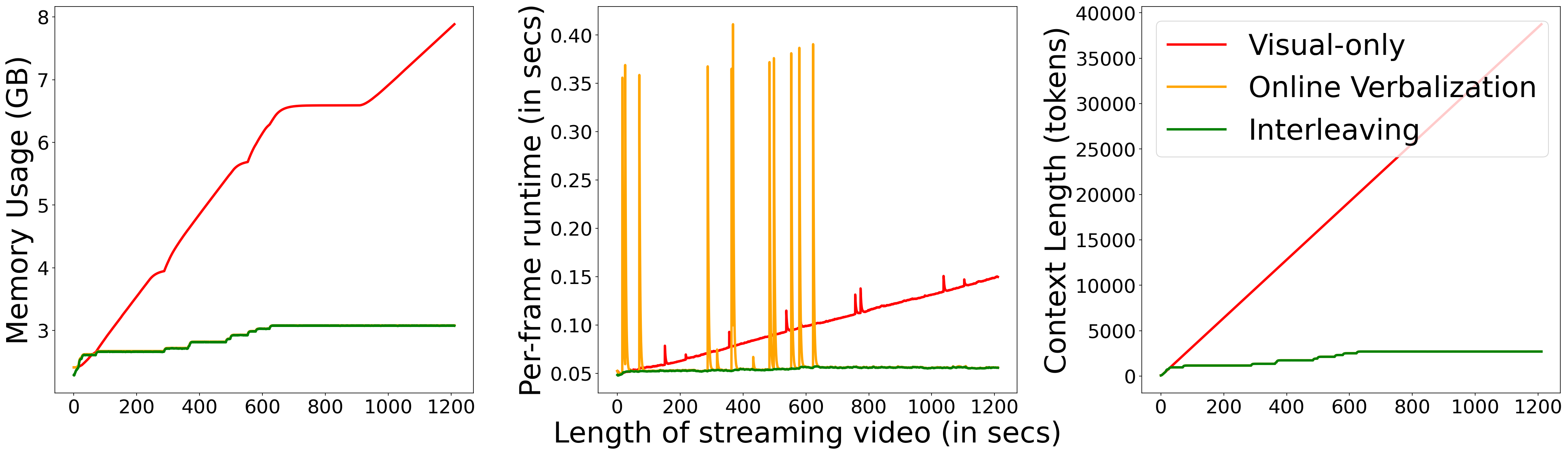}
\caption{\textbf{Scaling of memory, runtime and context length under various caching strategies.} Results reported on a subset of Ego4D-GoalStep online action detection validation set for a memory budget of 8GB on a single A6000 GPU. 
}
\label{fig::ablations:runtime}
\vspace{-0.6cm}
\end{figure}

\noindent\textbf{Verbalization \& Interleaving.} 
In~\cref{fig::ablations:runtime}, we analyze our verbalization and interleaving strategies using \mname-1B/5 on a subset of Ego4D Goal-Step, selecting videos spanning 15–30 minutes and sampling them at 4fps. We run the ablations on a fixed memory budget of 8GB on an A6000 GPU. Short-term cache length is set as $N_s = 64$ (16 seconds), and no limit is imposed on the long-term cache.
Using only visual tokens (red), memory scales linearly with video length, reaching 40K tokens to represent 20 minutes of video before exceeding 8GB memory. Such massive context also degrades performance.
Verbalization (orange) compresses 20 minutes of video into 2.5K tokens, reducing memory usage to 3GB and achieving sub-linear scaling. However, it introduces runtime instability, with frequent spikes in per-frame prediction time due to attention recomputation when converting short-term visual tokens into long-term text. Interleaving (green) in our multimodal cache mitigates these spikes, improving runtime stability while maintaining sub-linear scaling in both memory and compute.

\begin{table}[t] 
\centering
\setlength{\tabcolsep}{3mm}{
\resizebox{0.96\columnwidth}{!}{
\centering
\begin{tabular}{@{}ccccc@{}}
\toprule
\multirow{2}{*}{\textbf{\begin{tabular}[c]{@{}c@{}}{[}CLS{]}\\ Token\end{tabular}}} & \multirow{2}{*}{\textbf{\begin{tabular}[c]{@{}c@{}}Patch\\ Tokens\end{tabular}}} & \multirow{2}{*}{\textbf{\begin{tabular}[c]{@{}c@{}}\#Tokens\\ /frame\end{tabular}}} & \multicolumn{2}{c}{\textbf{Ego Val./ Acc. (\%)}} \\ \cmidrule(l){4-5} 
\multicolumn{1}{c}{} & & & \textbf{SigLIP} & \textbf{DINOv2} \\ \midrule
\rowcolor{blue!5}\multicolumn{5}{l}{w/~2-layer MLP~\cite{liu2024improved}} \\
\cmark & \xmark & 1 & 31.8$\pm$0.4 & 28.7$\pm$0.2 \\
\cmark & 16$\times$16 $\rightarrow$ 2$\times$2 & 5 & 32.1$\pm$0.6 & 32.4$\pm$0.5 \\
\cmark & 16$\times$16 $\rightarrow$ 4$\times$4 & 17 & 33.3$\pm$0.6 & 36.6$\pm$0.5 \\ \midrule
\rowcolor{blue!5}\multicolumn{5}{l}{w/~QFormer~\cite{li2023blip}} \\
\cmark & 16$\times$16 $\Rightarrow$ 4 & 5 & 32.9$\pm$0.6 & 33.6$\pm$0.5 \\
\cmark & 16$\times$16 $\Rightarrow$ 16 & 17 & 33.2$\pm$0.4 & 38.1$\pm$0.7 \\ \midrule
\rowcolor{blue!5}\multicolumn{5}{l}{w/~\cname{} (2 hands, 2 objects)} \\
\cmark & 16$\times$16 $\Rightarrow$ 0 & 5 & 30.9$\pm$0.3 & 34.7$\pm$0.5 \\
\cmark & 16$\times$16 $\Rightarrow$ 12 & 17 & 33.6$\pm$0.3 &40.7$\pm$0.6 \\ \bottomrule
\end{tabular}
}} 
\caption{\textbf{\cname{} ablations} on EgoExo4D fine-grained keysteps. Llama-3.2-1B-Instruct is used as the language decoder. Results reported over 3 runs. ($\rightarrow$ means avg-pooling, $\Rightarrow$ means Q-Former compression.)}
\vspace{-0.1cm}
\label{tab:ablation_hand-qformer}
\end{table}

\subsection{Runtime Analysis} \label{sec:exp::runtime}
In~\cref{tab:runtime}, we report the per-frame streaming inference and streaming dialogue~\cite{chen2024videollm} performance of \mname{} variants on a single A6000 GPU. Our lightest variant, \mname-1B/5, achieves over 10 FPS with a minimal 2GB memory footprint, making it ideal for low-latency applications. Meanwhile, \mname-8B/11 offer higher performance but require more compute. For most videos in this dataset, \mname-1B/5 consistently maintains over 10 FPS, sometimes even reaching 20 FPS when the video has significant no-action frames. For per-frame predictions, a key bottleneck for VideoLLMs is their autoregressive text generation, adding around 6 extra iterations, corresponding to 5.7 tokens/(activity step) on average for Ego4D-GoalStep. This issue is mitigated with streaming dialogue~\cite{chen2024videollm}, where the model is trained to remain silent and respond to user queries only when necessary. In this setting, \mname-1B/5 delivers real-time performance at 24.6 FPS.

\begin{table}[t]
\centering
\setlength{\tabcolsep}{0.3mm}{
\resizebox{1\columnwidth}{!}{
\centering
\begin{tabular}{llcccr}
\hline
\multirow{2}{*}{\textbf{\begin{tabular}[c]{@{}l@{}} \small{Streaming}\\ \small{Predictions}\end{tabular}}} &
 \multirow{2}{*}{\textbf{\textbf{Model}  }} &
 \multicolumn{3}{c}{\textbf{FPS ($\uparrow$)}} &
 \multirow{2}{*}{\textbf{\begin{tabular}[c]{@{}c@{}}Memory \\ (GB) ($\downarrow$)\end{tabular}}} \\ \cline{3-5}
 &
 &
 \textbf{\begin{tabular}[c]{@{}c@{}}Vision \\ Enc.\end{tabular}} &
 \textbf{\begin{tabular}[c]{@{}c@{}}Connector \\ + LLM\end{tabular}} &
 \textbf{\begin{tabular}[c]{@{}c@{}}Full \\ Model\end{tabular}} &
 \\ \hline
\multirow{2}{*}{\small{Per-frame}} & \small{ \mname-1B/5} & 74.8 & 10.4 & \textbf{9.1} & \textbf{2.0} \\
 & \small{ \mname-8B/11} & 38.0 & 4.2 & 3.5 & 16.2 \\ \hline
\multirow{2}{*}{\small{Dialogue~\cite{chen2024videollm}}} & \small{ \mname-1B/5 } & 74.8 & 34.2 & \textbf{24.6} & \textbf{2.2} \\
 & \small{ \mname-8B/11 } & 38.0 & 23.7 & 17.2 & 16.9 \\ \hline
\end{tabular}
}}
\caption{\textbf{Runtime Analysis} on a single A6000 GPU for the validation set of Ego4D-GoalStep online step detection for per-frame streaming and Ego4D-GoalStep Narration~\cite{wu2024videollm}. 
}\label{tab:runtime}
\end{table}

%%%%%%%%%%%%%%%%%%%%%%%%%%%%%%%%%%%%%%%%%%%%%%%%%%%%%%%%%%%%%%%%%%%%%%%%

%%% Section: Conclusion %%%%%%%%%%%%%%%%%%%%%%%%%%%%%%%%%%%%%%%%%%%%%%%%
\section{Conclusion and Limitations} 
In this paper, we presented~\mname{} for real-time understanding of procedural videos by interleaving two types of tokens in a vision-text multimodal cache -- verbalized text tokens, which provide compressed summaries of long-term observations, and visual tokens, which capture fine-grained details from short-term observations. We show that  \mname{}  reduces the token count for representing long-form procedural videos while effectively capturing critical spatio-temporal dependencies. We demonstrated state-of-the-art results on six tasks across four datasets, showing real-time per-frame inference at 10 FPS and  streaming dialogue at 25 FPS. \mname{} is the first real-time framework unifying recognition, anticipation, and planning for procedural assistance. Our multimodal interleaved cache is also generic, applicable to non-procedural videos, and unlocking a new area of compressing long  videos on the fly while reducing redundancy and boosting performance. \\
\noindent\textbf{Limitations.} DETR-QFormer, trained on pseudo ground-truth bounding boxes, generalizes well but could benefit from GT data to reduce biases from noisy annotations. Additionally~\mname{} focuses on detection but lacks real-time feedback on the \textit{``how''} of task execution. Finally, as discussed in \cref{sec:exp::generalization}, \mname's generalization performance remains modest, presenting a potential direction for future work.
%%%%%%%%%%%%%%%%%%%%%%%%%%%%%%%%%%%%%%%%%%%%%%%%%%%%%%%%%%%%%%%%%%%%%%%%

{
\small
\bibliographystyle{ieeenat_fullname}
\bibliography{main}
}

\end{document}

%% file: preamble.tex
\usepackage[dvipsnames]{xcolor}

\usepackage{microtype}
\usepackage{multicol, multirow, booktabs}
\usepackage{adjustbox}
\usepackage{appendix}
\usepackage{amsmath}
\usepackage{float}
\usepackage{graphicx}
\usepackage{colortbl}
\usepackage{soul}
\usepackage{pifont}
\usepackage{algorithm}
\usepackage{algpseudocode}
\usepackage{amsmath}
\usepackage{xcolor}

\newcommand{\mname}{ProVideLLM}
\newcommand{\cname}{DETR-QFormer}

\newcommand{\cmark}{\color{teal} \ding{51}}
\newcommand{\xmark}{\color{red} \ding{55}}

\definecolor{keywordcolor}{RGB}{0, 102, 204}
\definecolor{commentcolor}{RGB}{0, 153, 0}
\definecolor{functioncolor}{RGB}{153, 0, 153}
\definecolor{brightpink}{rgb}{1.0, 0.0, 0.5}

\newcommand\blfootnote[1]{%
  \begingroup
  \renewcommand\thefootnote{}%
  \footnote{#1}%
  \addtocounter{footnote}{-1}%
  \endgroup
}